\documentclass{article}
\usepackage{ijcai19}

\usepackage{times}
\usepackage{xcolor}
\usepackage{soul}
\usepackage[utf8]{inputenc}
\usepackage[small]{caption}

\usepackage{graphicx}        
\usepackage{mathtools}
\usepackage{bm}
\usepackage{makecell}
\usepackage{subcaption}
\usepackage{algorithm}
\usepackage[noend]{algpseudocode}
\usepackage{xcolor}

\newtheorem{definition}{Definition}
\newtheorem{lemma}{Lemma}

\graphicspath{{./figs/}}

\title{\LARGE \bf
Support Relation Analysis for Objects in Multiple View RGB-D Images
}

\author{Peng Zhang \and Xiaoyu Ge and Jochen Renz\\
Research School of Computer Science\\
        The Australian National University\\
        {\tt\small \{p.zhang, xiaoyu.ge, jochen.renz\}@anu.edu.au}%
}

\begin{document}
\maketitle
\thispagestyle{empty}
\pagestyle{empty}


\begin{abstract} 
Understanding physical relations between objects, especially their support relations, is crucial for robotic manipulation. There has been work on reasoning about support relations and structural stability of simple configurations in RGB-D images. In this paper, we propose a method for extracting more detailed physical knowledge from a set of RGB-D images taken from the same scene but from different views using qualitative reasoning and intuitive physical models. Rather than providing a simple contact relation graph and approximating stability over convex shapes, our method is able to provide a detailed supporting relation analysis based on a volumetric representation. Specifically, true supporting relations between objects~(e.g., if an object supports another object by touching it on the side or if the object above contributes to the stability of the object below) are identified.
We apply our method to real-world structures captured in warehouse scenarios and show our method works as desired.
\end{abstract}

\section{Introduction} \label{sec:1}
Scene understanding for RGB-D images has been extensively studied recently with the availability of affordable cameras with depth sensors such as Kinect~\cite{kinect}.
Among various scene understanding aspects~\cite{chen20163d}, understanding spatial and physical relations between objects is essential for robotics manipulation tasks~\cite{mojtahedzadeh2013automatic}, especially when the target object belongs to a complex structure, which is common in real-world scenes. Although most research on robotics manipulation and planning focuses on handling isolated objects~\cite{ciocarlie2014towards,kemp2007challenges}, increasing attention has been paid to the manipulation of physically connected objects~(for example~\cite{stoyanov2016no,li2017visual}). 
There are several problems that we need to deal with when analysing more complex object structures.
For example,  connected objects may remain stable due to support from adjacent objects rather than simple surface support from the bottom. 
In fact, support force may come from an arbitrary direction. Therefore, a simple bottom-up supporting relation ayalysis is not sufficient. 
Additionally, objects may hide behind other objects when observing from a certain view point. Given that real-world objects often have irregular shapes, correctly segmenting the objects and extracting their contact relations are challenging tasks. 
In order to solve these problems, an efficient physical model which deals with objects with arbitrary shapes is required to infer precise support relations of a structure.

In this paper, we propose a framework that takes raw RGB-D images as input and produces detailed support relations between objects in a stack. Most existing work on similar topics either assumes object shapes to be simple convex shapes~\cite{shao2014imagining}, such as cuboid and cylinder or makes use of previous knowledge of the objects in the scene~\cite{silberman2012indoor,song2016semantic} to simplify the support analysis process. Although reasonable experimental results were demonstrated, those methods usually lack the capability of dealing with scenes that contain a lot of unknown objects. As a significant difference to existing methods, our proposed method does not assume any knowledge about the objects in a scene. After individually segmenting point clouds of each view of the scene, our method builds a volumetric representation based on Octree~\cite{rensselaer1980octree} for each view with information about hidden voxels. The octree of the whole scene combined from the input views is then constructed using spatial reasoning about the objects. This process allows us to precisely register input point clouds from different views and provide a reliable contact graph integrating all views that can then be used for a proper support relation analysis. We adopt an intuitive physical model to determine the overall stability of the structure. By iteratively removing contact force between object pairs, we can infer supporters of each object and then build the support graph. To the best of our knowledge, this is the first work that is able to explain the object support relations from a physical perspective.

\section{Related Work}
\label{sec:2}


There has been work on scene understanding about support relations from a single view RGB-D image in both computer vision and robotics. In computer vision, scene understanding helps to produce more accurate detailed segmentation results. The work described in~\cite{jia20133d} applied an intuitive physical model to qualitatively infer support relations between objects and the experimental results showed the improvement of segmentation results on simple structures. In \cite{silberman2012indoor}, the types of objects in indoor scenes were determined by learning from a labeled data set to get more accurate support relations. The above-mentioned papers both took a single image as input which limited the choice of the physical model since a significant amount of hidden information was not available. \cite{shao2014imagining} attempted to recover unknown voxels from single view images by assuming the shape of the hidden objects to be cuboid and use static equilibrium to approximate the volume of the incomplete objects.
In robotics, \cite{mojtahedzadeh2013automatic}  proposed a method to safely de-stack boxes based on geometric reasoning and intuitive mechanics, which was shown to be effective in their later work~\cite{stoyanov2016no}. In \cite{li2017visual}, a simulation based method was proposed to infer stability during robotics manipulation on cuboid objects. This method includes a learning process using a large set of generated simulation scenes as a training set.

Humans look at things from different angles to gather comprehensive information for a better understanding. For example, before a jenga player takes an action, the player will usually look around the stack from several critical views to have an overall understanding of the scene. This also applies to robots when they take images as the input source. A single input image provides incomplete information. Even when the images are taken from different views of the same static scene, the information may still be inadequate for scene understanding when using quantitative models for inferring detailed physical and spatial information, as this requires precise input. Qualitative reasoning has been demonstrated to be more suitable for modeling incomplete knowledge~\cite{KUIPERS1989571}.
There are various qualitative calculi for representing different aspects of spatial entities~\cite{rcc,liu2009combining,ligozat1998reasoning,guesgen1989spatial,lee2013starvars}. One qualitative calculus that seems particularly useful for reasoning about spatial structures and their stability is the 
\emph{Extended Rectangle Algebra~(ERA)} \cite{zhang2014qualitative} which simplifies the idea in \cite{ge2013representation} to infer stability of 2D rectangular objects. It is possible to combine ERA with an extended version of \emph{cardinal direction relations} \cite{navarrete2006spatial} to qualitatively represent detailed spatial relations between objects, which helps to infer the transformation between two views. It is worth mentioning that \cite{panda2016single} proposed a framework to analyze support order of objects from multiple views of a static scene, yet this method requires relatively accurate image segmentation and the order of the images for object matching.

Models for predicting stability of a structure have been studied for many decades. Fahlman~\cite{fahlman1974planning} proposed a model to analyze system stability based on Newton's Laws. Simulation based models were also presented in recent years~\cite{cholewiak2013visual,li2017visual}. However, \cite{davis2016scope} argues that probabilistic simulation based methods are not suitable for automatic physical reasoning due to some limitations including the lack of capability to handle imprecise input. Thus in our approach, we aim to apply qualitative spatial reasoning to combine raw information from multiple views to extract understandable and more precise relations between objects in the environment.

\section{Method Pipeline}
\label{sec:3}

We now describe the overall pipeline of our support relation extraction method, which consists of three modules: image segmentation, view registration and stability analysis.

\begin{figure}[t]
\centering
\includegraphics[width=0.2\textwidth]{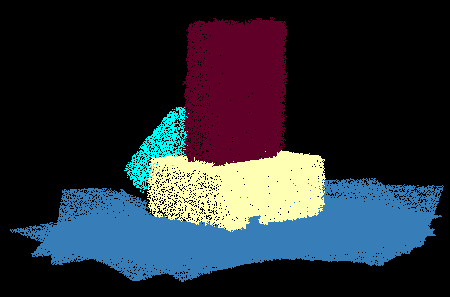}
\captionsetup{justification=centering}
\vspace{-2mm}
\caption{Segmentation of aligned images.}
\vspace{-4mm}
\label{fig:seg}
\end{figure}

\emph{The image segmentation module} takes a set of RGB-D images taken from different views of a static scene as input. To retain generality of our method, we do not assume any pre-known shapes of objects in the scene, that is, we do not use template matching methods that can provide more accurate segmentation results nor machine learning methods which require large amount of training data. This setting makes our method applicable in unknown environments. In the implementation, the raw rgbd data is first processed by a stream of morphology operations as described in \cite{ku2018defense} in order to fill the holes in the depth map. Notably, this hole-filling algorithm does not require any pre-training which is consistent with the no-prior-knowledge assumption in this paper. Then we use LCCP~\cite{lccp} for point cloud segmentation. LCCP first represents the point cloud as a set of connected supervoxels~\cite{papon2013voxel}. Then the supervoxels are segmented into larger regions by merging convexly connected supervoxels. Each point cloud of a view will be segmented into individual regions. We use a \emph{connected graph} to represent relations between the regions. Each graph node is a segmented region. The contact graph is then used to identify contact relation between objects in the structure. We use Manhattan world~\cite{furukawa2009manhattan} assumption to find the ground plane. The entire scene will then be rotated such that the ground plane is parallel to the flat plane. Details about segmentation and ground plane detection will not be discussed as we used this method with little change. Fig~\ref{fig:seg} shows a typical output from this module.

In the \emph{view registration module}, we use the iterative closest point~(ICP) algorithm~\cite{besl1992method} to find the transformation between two point clouds. Notably, the initial guess for ICP algorithm is crucial. A bad initial guess may lead the registration to a local minima which provides incorrect results~\cite{pomerleau2015review}. Due to the nature of multiple objects involving in the scene, we propose an algorithm to find an initial match for point clouds based on spatial relations between the objects. A matching between objects from two views will also be provided by this algorithm.
The contact graph of each single point cloud will then be combined to produce a contact relation graph over all input images after the registration of different views.



In the \emph{stability analysis} module, we adopt the definition of structural stability~\cite{livesley1978limit} and analyze static equilibrium of the structure by representing reacting forces at each contact area as a system of equations. A structure is considered stable if the equations have a solution. Given a static input scene, several schemes will be used to adjust the unseen part of the structure to make the static equilibrium hold.

The contribution of this paper is bi-fold. First, we introduce a qualitative reasoning method to extract spatial relations between objects in a stack. We apply this information to find proper initial guess of the ICP algorithm to demonstrate the its usefulness. Second, we propose a method for reconstructing volumetric model of objects with no prior knowledge about objects, which is then used to analyse the true support relation of the object stack.

\section{View Registration}
In this section, we introduce a qualitative spatial reasoning approach to match objects from two scenes in order to find a proper initial guess for ICP to register the point clouds. In subsection~\ref{sec:qsr}, the qualitative spatial calculi and definitions related to the initial guess estimation algorithm are introduced first. In subsection~\ref{sec:init_guess}, the algorithm is explained in detail.

\subsection{Preliminaries on Qualitative Spatial Reasoning}
\label{sec:qsr}




The \emph{extended rectangle algebra~(ERA)} \cite{zhang2014qualitative} is a qualitative spatial calculus which can be used to reason about the structural stability of connected 2D rectangular objects. 
For our problem, \emph{ERA} is not expressive enough as the objects are incomplete 3D entities with irregular shapes. In section~\ref{sec:3}, we mentioned that the ground plane has been detected under the Manhattan space assumption, thus it is possible to analyze spatial relations separately from vertical and horizontal directions. Although we do not assume all images be taken from the same height relative to the ground, it is reasonable to assume that images are taken from a human-eye view, not a birds-eye view. As the ground plane is detected, vertical spatial relations become stable to view changes. In contrast, horizontal spatial relations change dramatically when the view point changes. In order to analyze the horizontal spatial relations independently, all regions are projected onto the ground plane, i.e., a 2D Euclidean space.

\emph{ERA} relations can be represented using \emph{extended interval algebra~(EIA)} relations~(see table~\ref{tab:eia}) in each dimension in a 2D Euclidean space. EIA corresponds to Allen's interval algebra~\cite{allen1983planning} with an additional center point for each interval. As a result, \emph{EIA} has 27 basic relations~(denoted by $B_{eint}$) which produce $27^2$ \emph{ERA} relations~(see \cite{zhang2014qualitative} for formal definitions of \emph{ERA}). The $ERA$ relation for two regions $A$ and $B$ can be written as $ERA(A,B) = (EIA_x(A,B), EIA_y(A,B))$. We will infer changes of direction relations with respect to horizontal view changes by applying \emph{ERA}.

\begin{table}[h]
\vspace{-3mm}
\scriptsize
\begin{tabular}{|c|c|c|}
\hline
Relation & Illustration & Inverse Relation\\ \hline
\thead{$EIA(A,B)=lol$}&\includegraphics[width=0.12\textwidth,height = 0.2cm]{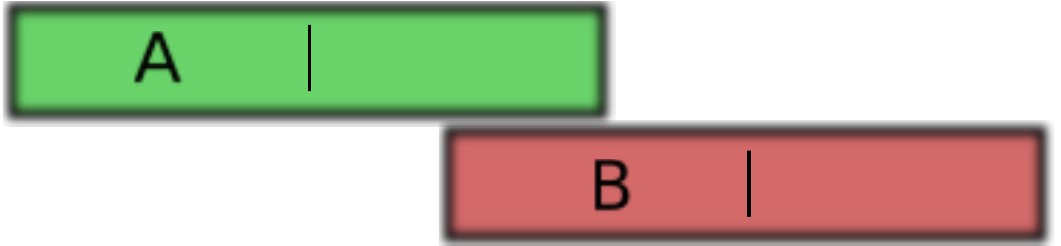}\centering &$EIA(B,A)=loli$\\ \hline
\thead{$EIA(A,B)=mol$}&\includegraphics[width=0.12\textwidth,height = 0.2cm]{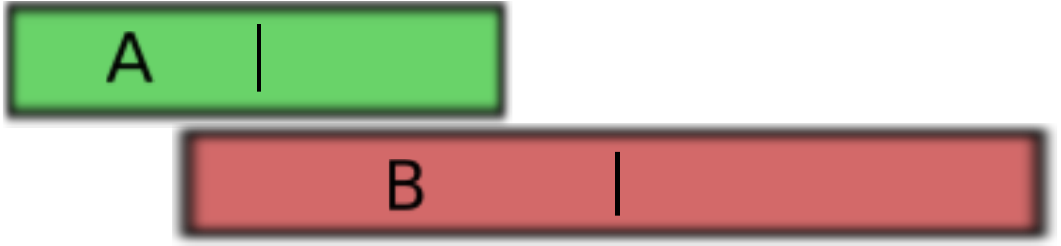}\centering & $EIA(B,A)=moli$\\ \hline
\thead{$EIA(A,B)=lom$}&\includegraphics[width=0.12\textwidth,height = 0.2cm]{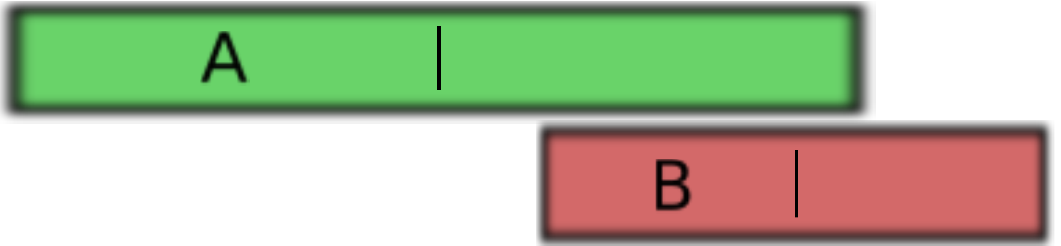}\centering &$EIA(B,A)=lomi$\\ \hline
\thead{$EIA(A,B)=mom$}&\thead{\includegraphics[width=0.12\textwidth,height = 0.2cm]{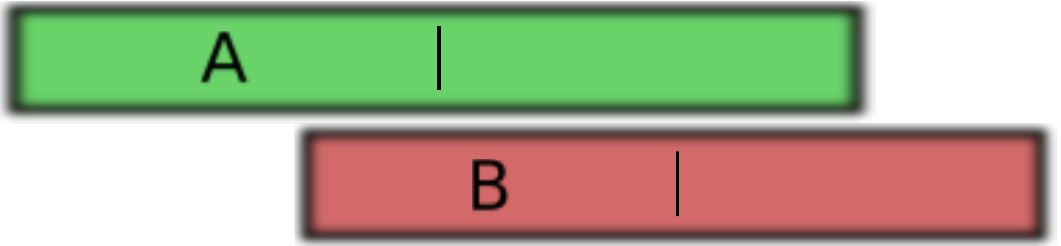}\centering }&$EIA(B,A)=momi$\\ \hline
\thead{$EIA(A,B)=ms$}&\thead{\includegraphics[width=0.12\textwidth,height = 0.2cm]{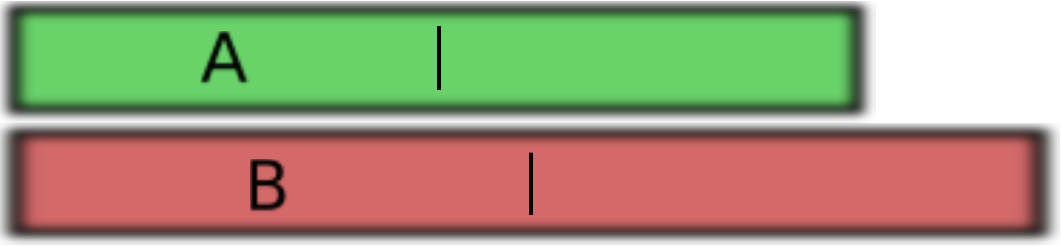}\centering }&$EIA(B,A)=msi$\\ \hline
\thead{$EIA(A,B)=ls$}&\thead{\includegraphics[width=0.12\textwidth,height = 0.2cm]{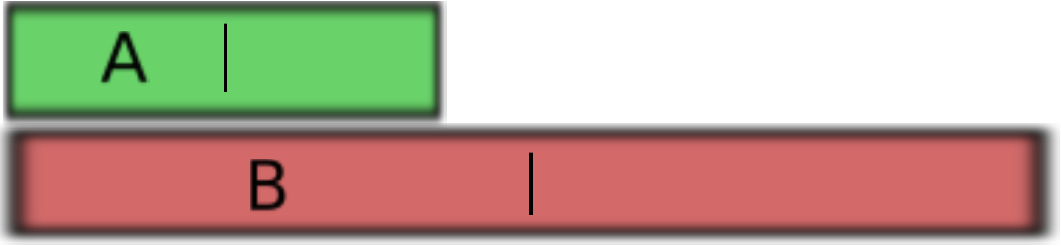}\centering }&$EIA(B,A)=lsi$\\ \hline
\thead{$EIA(A,B)=hd$}&\thead{\includegraphics[width=0.12\textwidth,height = 0.2cm]{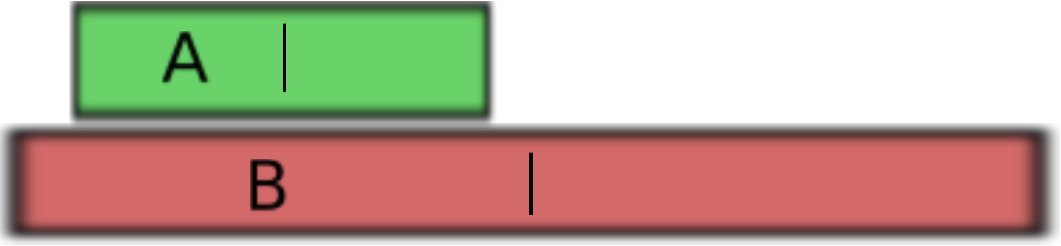}\centering }&$EIA(B,A)=hdi$\\ \hline
\thead{$EIA(A,B)=cd$}&\thead{\includegraphics[width=0.12\textwidth,height = 0.2cm]{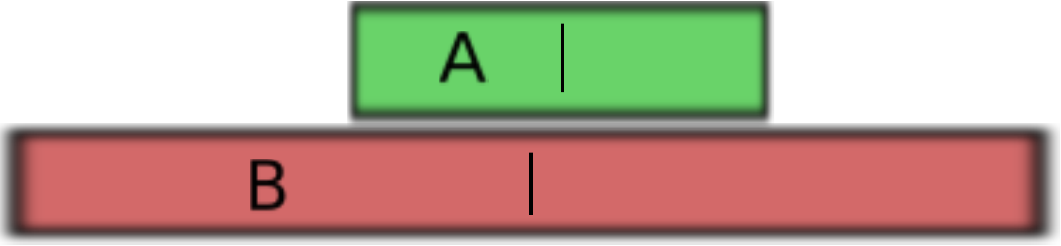}\centering }&$EIA(B,A)=cdi$\\ \hline
\end{tabular}
\vspace{-2mm}
\caption{Some example EIA relations (adding center point to IA)}
\vspace{-3mm}
\label{tab:eia}
\end{table}

\begin{definition}[region, region centroid, region radius]
Given a raw point cloud $PC$.  Region $a_l$ is the set of points ${p_1,...,p_n} \in PC$ with the same label $l$ from a segmentation algorithm.
Let $c$ = ($x_c$,$y_c$,$z_c$) denote the {\em region centroid}, where
\vspace{-4mm}
\begin{equation}
x_c = \frac{\underset{i=1}{\overset{n}{\Sigma}}x_{p_i}}{n},\quad
y_c = \frac{\underset{i=1}{\overset{n}{\Sigma}}y_{p_i}}{n},\quad
z_c = \frac{\underset{i=1}{\overset{n}{\Sigma}}z_{p_i}}{n}
\end{equation}
%
$dist(c,p_i)$ denotes the Euclidean distance between region centroid $c$ and an arbitrary point $p_i \in a_l$. The {\em region radius} $r$ of a region $a$ is:
\vspace{-3mm}
\begin{equation}
\quad \quad r = \underset{{i\in\{1,...,n\}}}{max}(dist(c,p_i))
\end{equation}

\end{definition}

\begin{figure}[t]
\centering
\includegraphics[width=0.4\textwidth, height = 2.5cm]{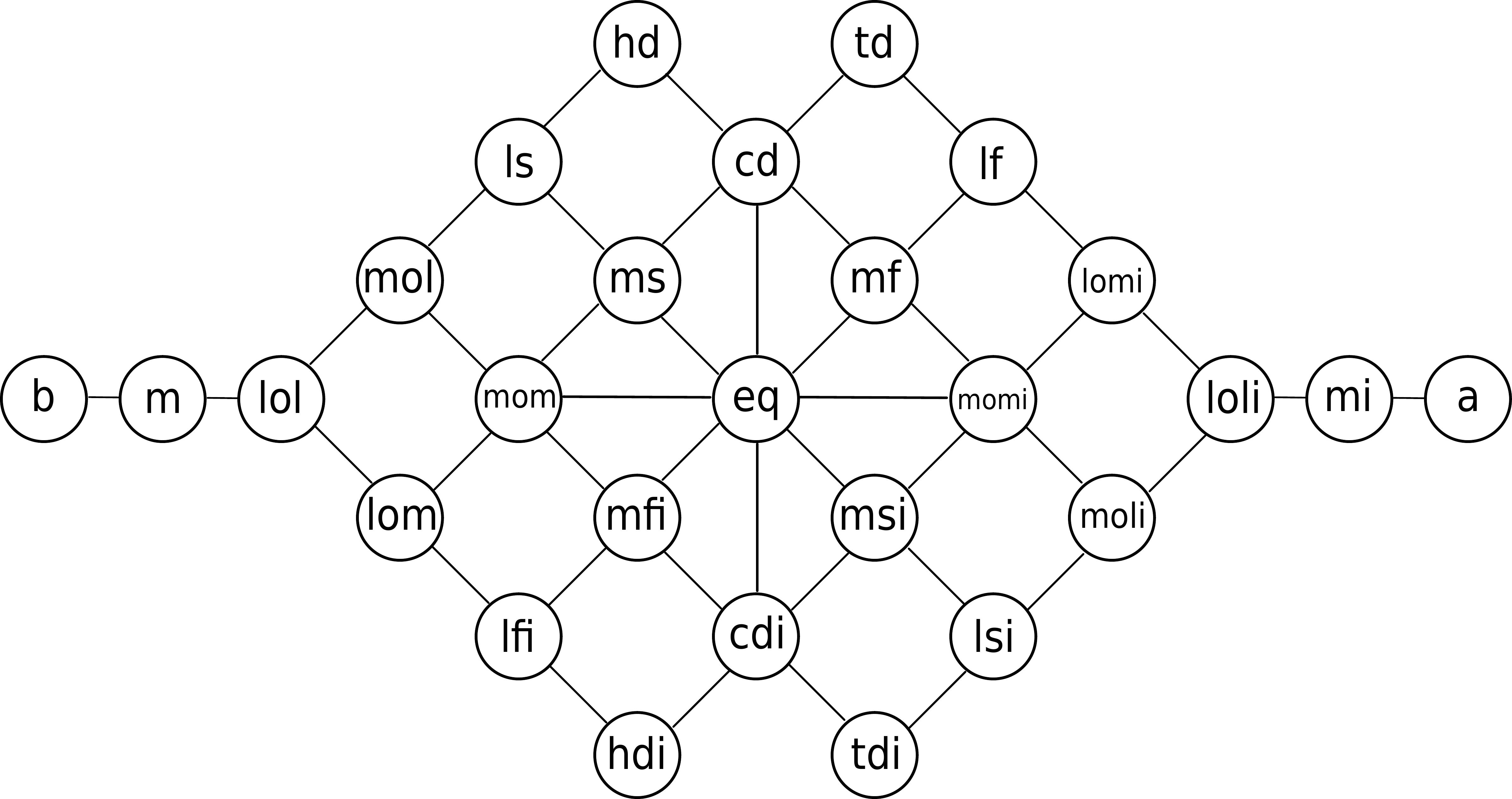}
\captionsetup{justification=centering}
\vspace{-2mm}
\caption{Conceptual neighborhood graph for $EIA$ based on horizontal view rotation}
\vspace{-4mm}
\label{fig:neighbor}
\end{figure}

Let $mbr$ denote the minimal bounding rectangle of a region. The $mbr$ will change with the change of views. As a result, the $EIA$ relation between two regions will change accordingly. By analyzing the $EIA$ change, an approximate horizontal rotation level can be determined between two views. Before looking at incomplete regions due to occlusion or noise from the sensor, we first research how the values of $ r_x^-, r_x^+, r_y^-, r_y^+$ change assuming the regions are completely sensed. We identify a conceptual neighborhood graph of $EIA$ which includes all possible one-step changes with respect to horizontal rotation of views~(see figure~\ref{fig:neighbor}).

\begin{definition}[view point, change of view]
The {\em view point} $v$ is the position of the camera.
Let $v_1$ and $v_2$ denote two view points, and $c$ be the region centroid of the sensed connected regions excluding the ground plane. Assuming the point cloud has been rotated such that the ground plane is parallel to the plane defined by x-axis and y-axis of a 3D coordination system. Let $v_{1xy}$, $v_{2xy}$ and $c_{xy}$ be the vertical projection of $v_1$, $v_2$ and $c$ to the xy plane.
The {\em change of view} $C$ is the angle difference between the line segments $c_{xy}v_{1xy}$ and $c_{xy}v_{2xy}$
\end{definition}

\begin{definition}[symmetric EIA relation]
Let $R \in B_{eint}$ to be an arbitrary EIA atomic relation. The symmetric EIA relation of $R$~(denoted by $symm(R)$) is defined as $R$'s axially symmetric atomic relation against the axis of symmetry formed by relations \{$cd,~ eq,~ cdi$\} in the conceptual neighborhood graph given in figure~\ref{fig:neighbor}.
For example, $symm(mol) = lomi$.
The symmetric relation of $cd$, $eq$ and  $cdi$ are themselves.
\end{definition}

\begin{lemma}
\label{lem::erachange}
Let $C_{cw\pi/2}$ denote the view change of $\pi/2$ clockwise from view point $v_1$ to $v_2$. Let $ERA_{ab_1} = (r_{x1},r_{y1})$ and $ERA_{ab_2}  = (r_{x2},r_{y2})$ denote the $ERA$ relations between region $a$ and $b$ at $v_1$ and $v_2$. \\Assuming the connected regions are fully sensed, then $r_{x2} = symm(r_{y1})$, $r_{y2} = r_{x1}$
Similarly, if the view changes by $\pi/2$ anticlockwise, then $r_{x2} = r_{y1}$, $r_{y2} = symm(r_{x1})$
\end{lemma}

Proof:
Lemma~\ref{lem::erachange} can be simply proved by reconstructing a coordination system at each view point.

Although the conceptual neighborhood graph indicates possible relation changing path for a pair of objects in one dimension, the way the changes happen depends on the rotation direction~(clockwise or anti-clockwise) and their $ERA$ relation before the rotation. For example, $ERA(A,B) = (m,m)$ means $mbr(A)$ connects $mbr(B)$ at the bottom-left corner of $mbr(B)$, thus if the view rotates anti-clockwise, $mbr(A)$ tends to move upwards related to $mbr(B)$ regardless of the real shape of $A$ and $B$, therefore $EIA_y(A,B)$ will change from `$m$' to `$lol$' but not `$b$'. To determine the changing trend of $ERA$ relations more efficiently, we combine $ERA$ with \emph{cardinal direction relations~(CDR)} which  describes how one region is relative to the other in terms of directional position.

\begin{figure}[t]
    \centering
\vspace{-2mm}
    \begin{subfigure}[b]{0.15\textwidth}
    \centering
        \includegraphics[width=\textwidth]{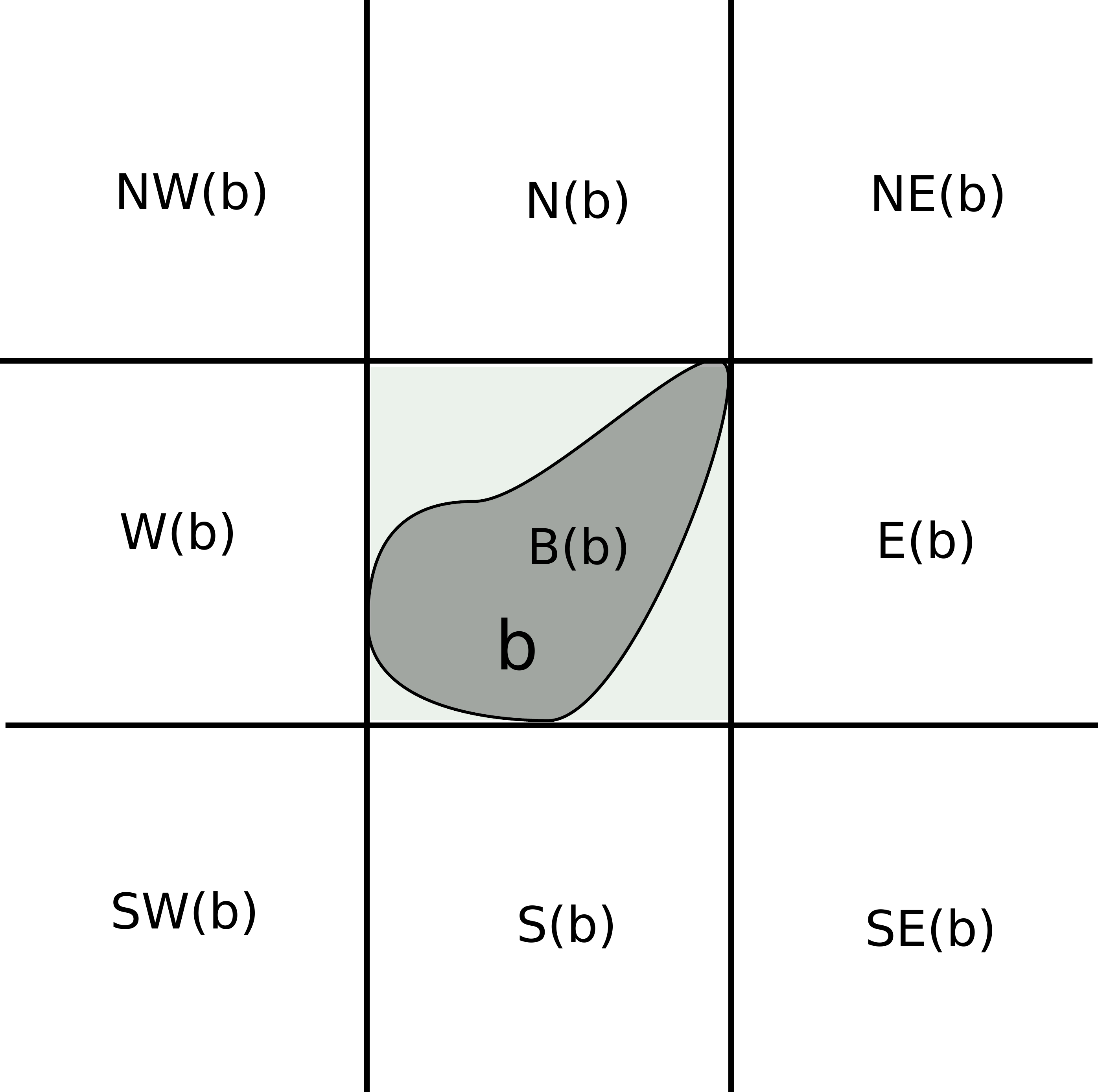}
        \label{fig:cdr}
    \end{subfigure}
    \qquad
    ~ 
    \begin{subfigure}[b]{0.15\textwidth}
    \centering
        \includegraphics[width=\textwidth]{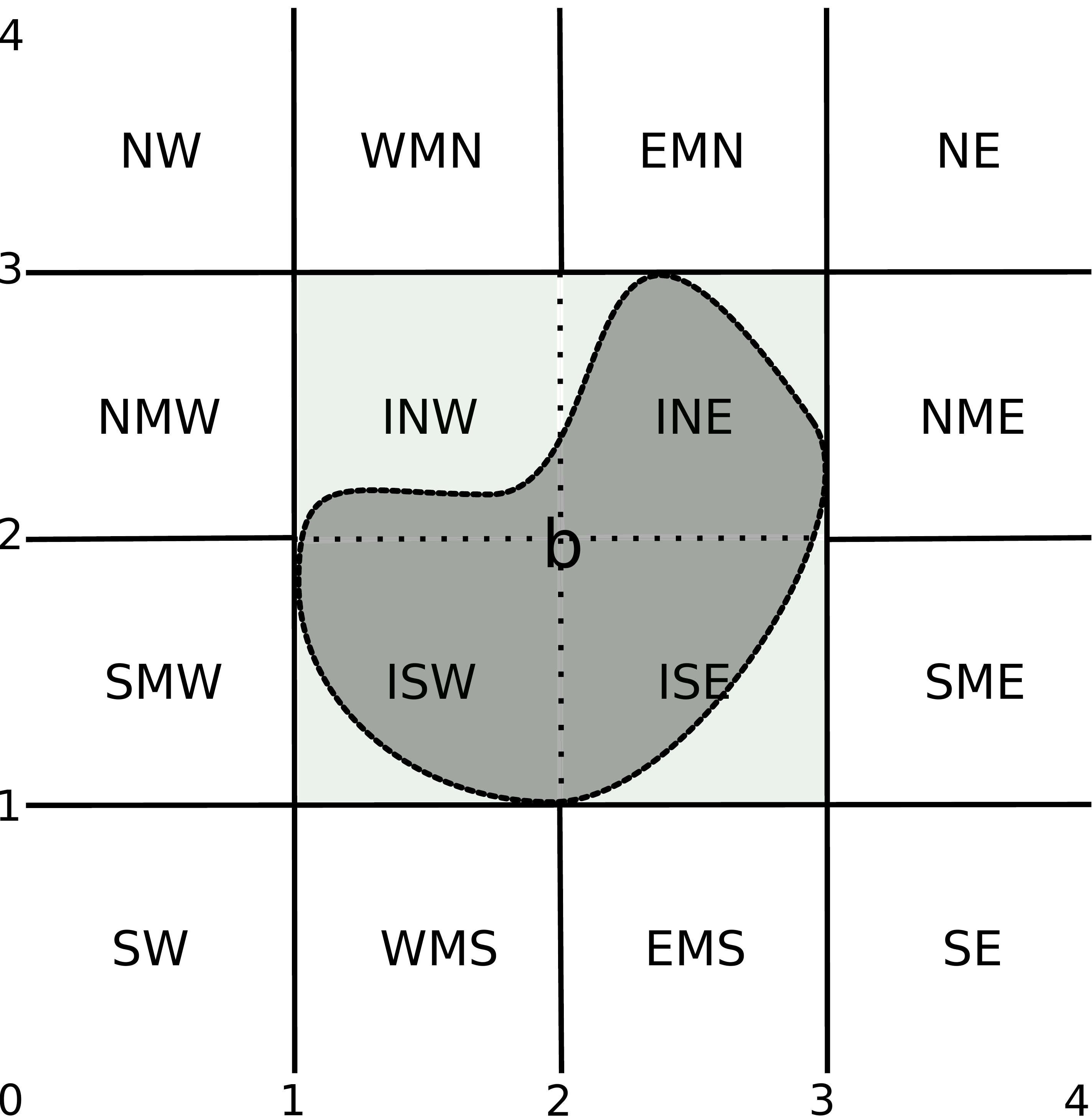}
    \label{fig:ecdr}
    \end{subfigure}
\vspace{-5mm}
     \caption{Basic CDR (left) and Extended CDR~(ECDR) (right).}\label{fig:cdr_all}
\vspace{-3mm}
\end{figure}

The basic $CDR$ \cite{skiadopoulos2004composing} contains nine cardinal tiles as shown in figure~\ref{fig:cdr_all}. `$B$' represents the relation `$belong$', the other eight relations the cardinal directions N (north), NE (north-east), etc.  

\begin{definition}[basic CDR relation]
\label{def:basiccdr}
A basic CDR relation is an expression $R_1:...:R_k$ with $1 \leq k \leq 9$ where:
\begin{enumerate}
\item $R_1,...,R_k \in \{B,N,NE,E,SE,S,SW,W,NW\}$
\item $R_i \neq R_j,~\forall 1 \leq i,j \leq k, i \neq j$
\item $\forall b \in REG,~\exists a_1,...,a_k \in REG~and ~ a_1 \cup ... \cup a_k \in REG$~(Regions that are homeomorphic to the closed unit disk ${(x, y): x^2 + y^2 \leq 1}$ are denoted by REG).
\end{enumerate}
If $k = 1$, the relation is called a single-tile relation and otherwise a multi-tile relation.
\end{definition}

Similar to the extension from $RA$ to $ERA$, we introduce center points to extend basic $CDR$ to the extended $CDR$~(denoted as $ECDR$, see figure~\ref{fig:cdr_all}) in order to express inner relations and detailed outer relations between regions.
Notably, a similar extension about inner relations of CDR was proposed in~\cite{liu2005internal}. However, as we focus on $mbr$ of regions in this problem, their notions for inner relations with trapezoids are not suitable for our representation.

In~\cite{navarrete2006spatial}, \emph{rectangular cardinal direction relations~(RCDR)} which combines $RA$ and $CDR$ was studied. As a subset of CDR, RCDR considers single-tile relations and a subset of multi-tile relations that represent relations between two rectangles whose edges are parallel to the two axes. We combine $ERA$ and $ECDR$ in similar way to produce the \emph{extended rectangular cardinal direction relations~(ERCDR)}. Including all single-tile and multi-tile relations, there exists 100 valid relations~(not all listed in this paper) to represent the directional relation between two $mbrs$.




\subsection{Initial Guess Estimation for ICP}
\label{sec:init_guess}
In this section, the algorithm for matching objects between two views is proposed. In section~\ref{sec:3}, the point cloud has been aligned to the direction of the ground, therefore, the only two factors for initial transformation estimation are rotation against the vertical axis and the translation. With the matched objects, the task for estimating initial transformation between two point clouds for ICP algorithm is then minimising the sum of distance between all matched object pairs.

\begin{figure}[t]
\centering
\includegraphics[scale=.2]{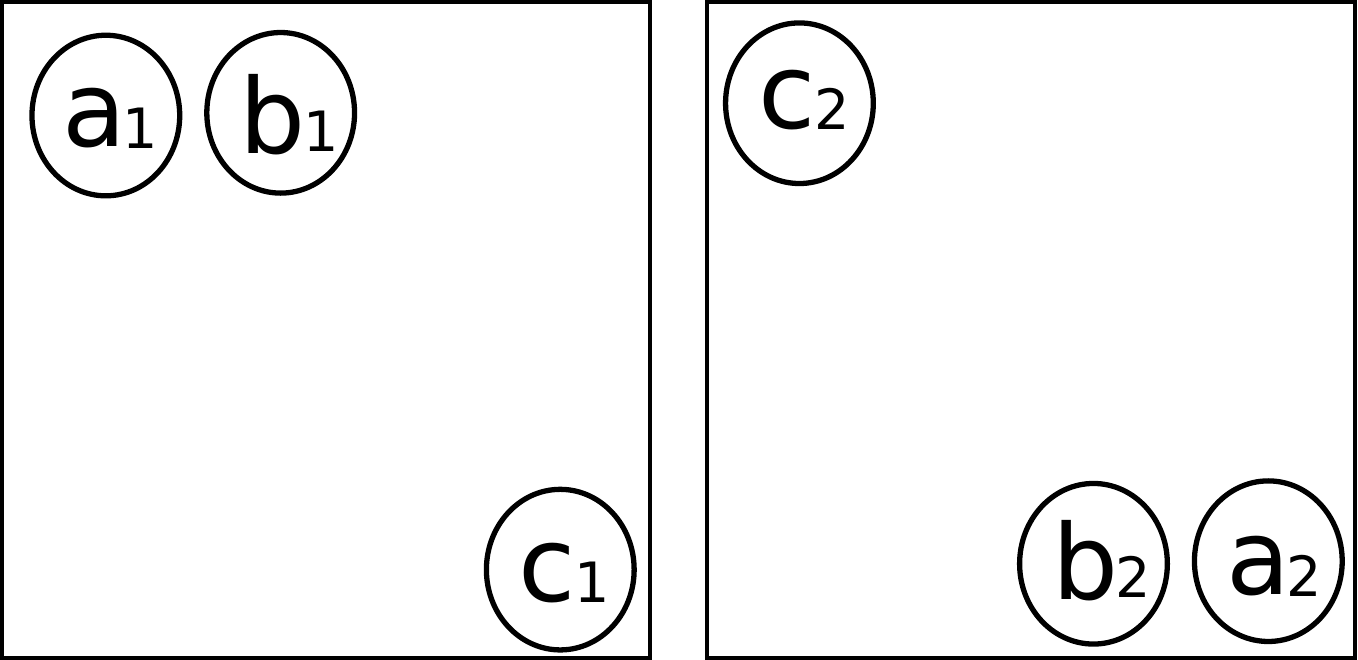}
%
%
\captionsetup{justification=centering}
\vspace{-2mm}
\caption{An example to illustrate how to use spatial reasoning to verify object match}
\label{fig:qsr_example}
\end{figure}

Informally, assuming a spatial relation graph is built for any two objects in the same view. If most objects in one view are correctly matched to the corresponding ones in the other view, the two spatial relation graph can be very similar~(if not identical because of incomplete input) by rotating one view by a certain angle. For example, in figure~\ref{fig:qsr_example}, the spatial relation graph for the left view is \{west($a_1$, $b_1$), northwest($a_1$, $c_1$), northwest($a_1$, $c_1$)\}, and for the right view is \{east($a_2$, $b_2$), southeast($a_2$, $c_2$), southeast($a_2$, $c_2$)\}. If correctly matching all three pairs of objects, the identical graph can be obtained by rotating the right view by $\pi$/2 from any direction. If wrongly matching any objects, the identical graph can never be obtained.

Therefore, a distance function is necessary for measuring the similarity of two relation graphs. A method about how to calculate the distance between two ERCDR relations is shown below.





\begin{definition}[directional property of a single-tile relation]
Each tile $t$ in ECDR has both a horizontal directional property~$HDP(t) \in \{E,W\}$ and a vertical directional property~$VDP(t) \in \{N,S\}$. The value is determined by the relative directional relation between the centroid of $t$ and the centroid of the reference region $b$. For example, $HDP(INE) = E$ and $VDP(INE) = N$.
\end{definition}

\begin{definition}[directional property of multi-tile relation]
The directional property of a multi-tile relation $mt$ is determined by majority single tile directional properties in the multi-tile relation, where $HDP(mt)\in \{E,M,W\}$ and $VDP(mt)\in \{N,M,S\}$, where $M$ represents `middle' which appears when the counts of the single tile directional properties are equal. For example, $HDP(WMN:EMN:INW:INE) = M$ and $VDP(WMN:EMN:INW:INE) = N$.
\end{definition}

Directional property can be used to estimate the trend of the relation change. Let a view point rotate clockwise, if HDP and VDP are as observed, then the  change trend is as described in the following table: 

\begin{tabular}{c|c|c|c}
HDP & VDP & change trend & change direction \\
\hline
$E$ & $S$ or $M$ & south to north & vertical \\
$E$ or $M$ & $N$ & east to west & horizontal \\
$W$ & $N$ or $M$ & north to south & vertical\\
$W$ or $M$ & $S$ & west to east & horizontal \\
\end{tabular}
%

One ERCDR relation can be transformed to the other by lifting the four bounding lines of the single/multi-tile. The distance $d$ between two ERCDR relations is calculated from horizontal and vertical directions by counting how many grids each boundary line lifts over. The distance is related to the direction of the view point changes as well as the directional properties of the region.
However, with the same angle difference, the inner tiles take much fewer changes than the outside tiles. We introduce the \emph{quarter distance size} to represent the angle change of $\pi/2$ for normalizing the distance between ERCDR relation pairs corresponding to the angle difference.

Based on lemma~\ref{lem::erachange}, we can infer the ERA relation between two regions after rotating the view point by $\pi/2$, $\pi$ and $3\pi/2$ either clockwise or anticlockwise. The ERA relation can be then represented by an ERCDR tile.
In order to calculate the distance between ERCDR tiles, we label each corner of the single tiles with a 2D coordinate with bottom-left corner of tile $SW$ to be the origin $(0,0)$~(see figure~\ref{fig:cdr_all}).

\begin{definition}[distance between ERCDR tiles]
Let $t_1$ and $t_2$ be two ERCDR tiles. $x_1^-$ and $x_1^+$ are the left and right bounding lines of $t_1$,  $y_1^-$ and $y_1^+$ the top and bottom bounding lines of $t_1$. $x_2^-$ and $x_2^+$ are the left and right bounding lines of $t_2$. $y_2^-$ and $y_2^+$ the top and bottom bounding lines of $t_2$.

The unsigned distance between   $t_1$ and $t_2$ is calculated as: \vspace{-2mm}
\begin{equation}
|d(t_1,t_2)| = |(x_2^+ - x_1^+) + (x_2^- - x_1^-)| + |(y_2^+ - y_1^+) + (y_2^- - y_1^-)|
\end{equation}

The sign of $d$ is determined by whether the change trend suggested by directional property is followed. If so, $d(t_1,t_2)$ = $|d(t_1,t_2)|$, else $d(t_1,t_2)$ = $-|d(t_1,t_2)|$. If there is no or symmetric change on the trend direction, $d(t_1,t_2) = 0$
\end{definition}

If there is a significant angle difference between the two tiles, the distance may not be accurate due to multiple path for the transformation. Here we introduce three more reference tiles by rotating the original tile by $\pi/2$, $\pi$ and $3\pi/2$ in turn.

\begin{definition}[quarter distance]
Let $t_1$ be an ERCDR tile and $t_1'$ be the ERCDR tile produced by rotating $t_i$ by $\pi/2$ or $-\pi/2$. 
By applying lemma~\ref{lem::erachange}, the reference tiles can be easily mapped to ERCDR tiles. The quarter distance of $t_1$ is defined as:
\vspace{-2mm}
\begin{equation}
d_{qt} = d(t_1,t_1')
\end{equation}
\end{definition}

\begin{definition}[normalized distance]
Let $t_1$ be an ERCDR tile. $t_{\pi/2}$, $t_{\pi}$ and $t_{3\pi/2}$ denote the three reference tiles for $t_1$. Let $t_2$ be another ERCDR tile. The {\em normalized distance} $d_{norm}(t_1,t_2)$ will be calculated in two parts:
\begin{enumerate}
\item The base distance $d_{base}$.
Let $T = \{t_1,t_{\pi/2},t_{\pi},t_{3\pi/2} \}$.
\begin{equation}
d_{base}(t_1,t_2) = \begin{cases}
  0 \quad if \quad \underset{t \in T}{argmin}(|d(t,t_2)|) = t_1\\
  1 \quad if \quad \underset{t \in T}{argmin}(|d(t,t_2)|) = t_{\pi/2}\\
  2 \quad if \quad \underset{t \in T}{argmin}(|d(t,t_2)|) = t_{\pi}\\
  3 \quad if \quad \underset{t \in T}{argmin}(|d(t,t_2)|) = t_{3\pi/2}
\end{cases}
\end{equation}
%

\item The normalized distance $d_{norm}(t_1,t_2) = d_{base}(t_1,t_2) +$ \\ \vspace{-2mm}
\begin{equation}
\begin{split}
 \frac{d(\underset{t \in T}{argmin}(|d(t,t_2)|),t_2)+1}{d_{qt}(t_1)+1}
\end{split}
\end{equation}

\end{enumerate}
\end{definition}

Having the normalized distance for calculating the similarity between two ERCDR tiles, we now show the algorithm for identifying proper matching between objects from two different views in algorithm~\ref{algo:match}.

\begin{algorithm}[!h]
\caption{Object Matching}\label{algo:match}
\begin{algorithmic}

\Function{GetPermutation}{}
\State Input: 

	$objIDList$,~ 
	$processedObjs$. 
	
	$length$ // number of objs in one list, equals to the smaller 				size of the two object list
	
\State $size~\leftarrow~objIDList.size()$

\If{$length == 1$}
	\For{$i \leftarrow 0;~i < size;~i++$}
	\State $temList~\leftarrow~processedObjs$
	\State $temList.pushback(objIDList[i])$
	\State $permutation.pushback(temList)$
	\EndFor
\Else	
	\For{$i \leftarrow 0;~i < size;~i++$}
	\State $temList~\leftarrow~processedObjs$
	\State $temList.pushback(objIDList[i])$
	\State $remainObjIDList~\leftarrow~objIDList$
	\State $remainObjIDList.erase(i)$
	\State $GetPermutation(remainObjIDList,$ 
	
			$temList, length-1, permutation)$
	\EndFor
\EndIf
	
\State Output: $permutation$
\EndFunction

\Function{GetMatchedObjects}{}
\State Input: 

		$objIDList1,~objIDList2$ 
		
		$relationGraph1, relationGraph2$ 
\State Output: $match$ // A set of matched object pairs 

\State //Assume objIDList1.size() $>$ objIDList1.size() 		
\State $size~\leftarrow~objIDList2.size()$
\State $emptyList~\leftarrow~\{\}$,~$permutation~\leftarrow~\{\}$
\State $GetPermutation(objIDList1,emptyList,$
		
		$ size, permutation)$
\State $candidateMatch~\leftarrow~\{\}$		
\For{$i \leftarrow 0;~i < permutation.size();~i++$}
	\State $temList~\leftarrow~\{\}$,~$curList~\leftarrow~permutation[i]$
	\For{$j \leftarrow 0;~j < objIDList2.size();~j++$}
		\State $temList.pushback({curList[j],objIDList2[j]})$
	\EndFor
	\State $candidateMatch.pushback(temList)$
\EndFor	

\State $err\leftarrow INFINITY$

\For{$i \leftarrow 0;~i < candidateMatch.size();~i++$}
	\State $curList~\leftarrow~candidateMatch[i]$,~$distList~\leftarrow~\{\}$
	\For{$m \leftarrow 0;~m < curList.size();~m++$}
		\For{$n \leftarrow 0;~n < curList.size();~n++$}
			\If{$m==n$}
				\State continue
			\EndIf
			\State $r1\leftarrow relationGraph1(curList[m][0],curList[n][0]) $
			\State $r2\leftarrow relationGraph2(curList[m][1],curList[n][1]) $

			\State $distList.pushback(d_{norm}(r1,r2))$
		\EndFor
	\EndFor
	\If {$Variance(distList) < err$}
		\State $err ~\leftarrow~ Variance(distList)$
		\State $match~\leftarrow~ curList$
	\EndIf 
\EndFor	
\Return $match$	
\EndFunction

\end{algorithmic}
\end{algorithm}

Once the match of objects has been determined, the initial transformation can be calculated by performing local search of horizontal rotation to obtain a minimal sum of distance between two relation graphs. Than a translation is also calculated by minimising the euclidean distance of the geometric centers of the matched objects. The ICP algorithm will be performed using the calculated initial transformation.



\section{Stability Analysis}
\label{stability}
In this section, we introduce a method to complete the object with invisible voxels. Then, we show how to get support relation from the volumetric representation of the objects.
\vspace{-2mm}
\subsection{Object Completion}
\begin{figure}[t]
\centering
\includegraphics[scale=.32]{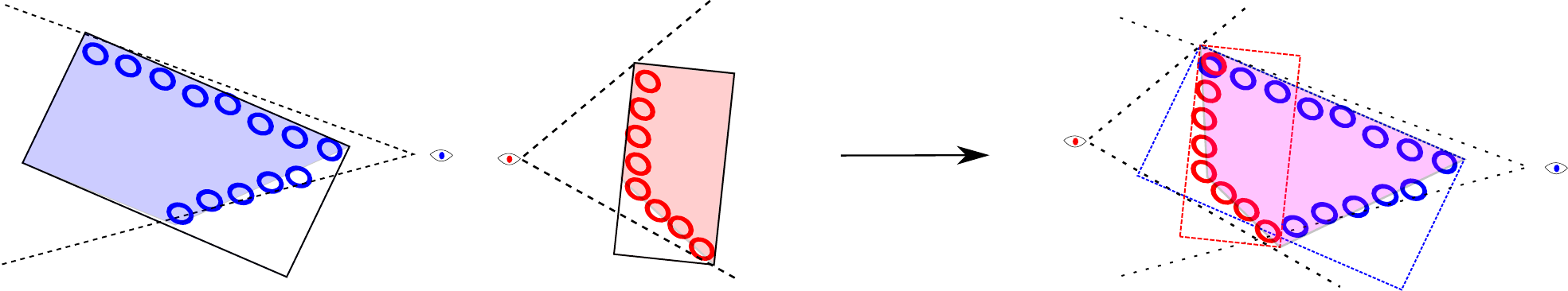}
%
%
\captionsetup{justification=centering}
\vspace{-2mm}
\caption{Object completion.}
\label{fig:complete}
 \vspace{-4mm}
\end{figure}

With the registration of muptiple views, an octree representation of the scene is built. The next step is to classify invisible voxels to the objects in order to analyse the support relation. First, for each of the object, an oriented minimal bounding box~{OMBB} is calculated. We perform Ransac~\cite{fischler1981random} algorithm to fit the largest plane to the object point cloud. This plane is used as one surface and the OMBB of the object is them determined.
All invisible voxels in the OMBB are then assigned to this object. As the octree is built with point clouds from different views, the set of invisible voxels are largely eliminated and the volumetric model tends to represents the intrinsic shape of the object. Figure~\ref{fig:complete} shows the process of object completion in 2D for simplicity.

 \vspace{-2mm}
\subsection{Support Relation Analysis}
We use a modified version of the structural analysis method in \cite{stacking}. A structure is in static equilibrium when the net force and net torque of the structure equal to zero. The static equilibrium is expressed in a system of linear equations \cite{whiting2009procedural}: \vspace{-4mm}
\begin{equation}\label{eq:staticEq}
\begin{array}{ccc}
\quad \quad \quad \quad  \quad \quad \bm{A_{eq}} \cdot \bm{f} + \bm{w} = \bm{0}\\
\quad \quad \quad \quad  \quad \quad \|\bm{f^n}\| \geq 0 & 1) \\
\quad \quad \quad \quad  \quad \quad  \|\bm{f^s}\| \leq \mu \|\bm{f^n}\| & 2)

\end{array}
\end{equation}
$A_{eq}$ is the coefficient matrix where each column stores the unit direction vectors of the forces and the torque at a contact point. To identify the contact points between two contacting objects, we first fit a plane to all points of the connected regions between the objects. We then project all points to the plane and obtain the minimum oriented bounding rectangle of the points. The resulting bounding rectangle approximates the region of contact, and the four corners of the rectangle will be used as contact points.  $f$ is a vector of unknowns representing the magnitude of each force at the corresponding contact vertex. The forces include contact forces $\bm{f^n}$ and friction forces $\bm{f^s}$ at the contact vertex. The constraint 1) requires the normal forces to be positive and constraint 2) requires the friction forces comply with the Coulomb model where $\mu$ is the coefficient of static friction. A structure is stable when there is a solution to the equations.

Using the structural analysis method, we can identify support relations between objects. Specifically, we are interested in identifying the core supporters \cite{ge2016visual} of each object in a scene. An object $o_1$ is a core supporter of another object $o_2$ if $o_2$ becomes unstable after removal of $o_1$. Given a contact between $o_1$ and $o_2$, to test whether $o_1$ is the core supporter of $o_2$, we first identify the direction vectors of forces and torque given by the contact on $o_2$, and set them to zero in Eq.~\ref{eq:staticEq}. This is equivalent to removing all forces that $o_1$ imposes on $o_2$. If the resulting Eq.~\ref{eq:staticEq} has no solution, then $o_1$ is the core supporter. We test each pair of objects in a scene and obtain a support graph, which is defined as a directed graph with each vertex representing an object. There is an edge from $v_1$ to $v_2$ if $o_1$ is a core supporter of $o_2$.

\section{Experiments}
We first test our method about estimation of initial guess for ICP algorithm. Then we show the method's capability of identifying core supporters of an object in a structure. For both experiments, we test our method on two data sets as well as some single scenes for testing special configurations. \emph{Data set 1} is from~\cite{panda2016single} which contains seven different scenes. \emph{Data set 2} is taken from a warehouse scenario of a real logistics application setting in sorting parcels. This data set consists of 5 scenes.

\subsection{Initial Guess Estimation of ICP}

\begin{figure}[t]
\centering
\includegraphics[scale=.25]{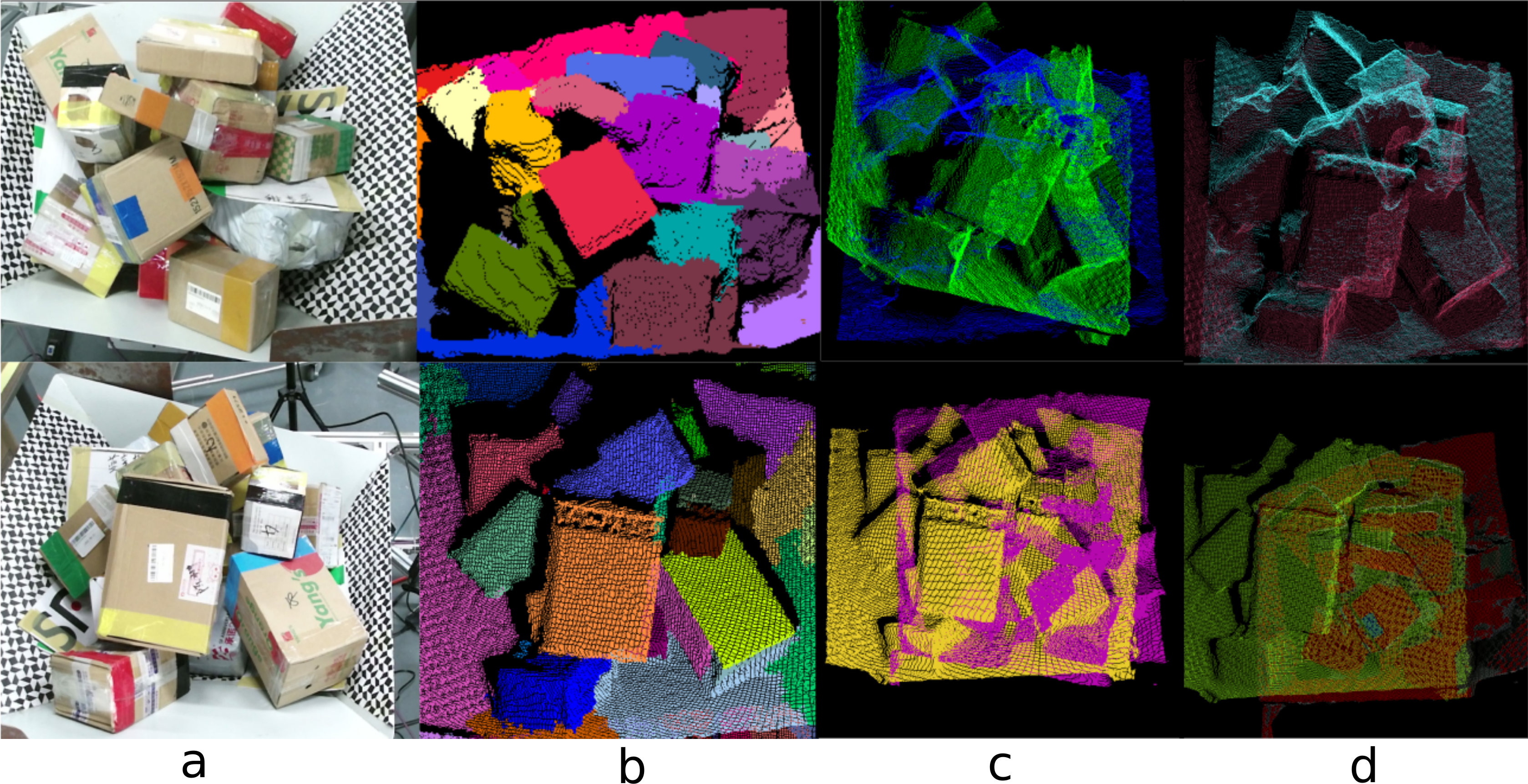}

\captionsetup{justification=centering}
\vspace{-2mm}
\caption{Results of ICP initial guess estimation~(data set 2).\\
Col. a shows the photos of different views; col. b is the segmentation of single views; the top image of col. c is the initial guess from algorithm\ref{algo:match}; the bottom is the random initial guess; the top image of col. d is the registration result with initial guess from algorithm~\ref{algo:match}; the bottom one is from random initial guess.}
\label{fig:align}
\end{figure}

In this experiment, we compare the initial guess from algorithm~\ref{algo:match} with random initial guess for ICP point cloud registration. Figure~\ref{fig:align} shows the result qualitatively. In table~\ref{tab:guess}, we use the mean sum of squared error (MSE) for all point pairs to evaluate the quality of the registration on data set 2 which consists of more complex scenes.

\begin{table}
\scriptsize
\centering
\begin{tabular}{|c|c|c|c|}
\hline
	&MSE (Data set 2)\\
\hline
Algorithm~\ref{algo:match}&1.132e-3\\
\hline
Random initial guess&2.29e-3\\
\hline
\end{tabular}
\caption{Initial guess results}
\label{tab:guess}
 \vspace{-5mm}
\end{table}

\subsection{Support Graph Evaluation}

For core supporter detection, we show that our method out performs the method in~\cite{panda2016single} on data set 1. As the method in~\cite{panda2016single} requires precise object models for segmentation, it does not work in unknown scenarios in data set 2. Therefore, only algorithm~\ref{algo:match} is tested on data set 2. The reason why support relation accuracy is slightly low is that there are more errors from segmentation which provide more false positive support relations.

\begin{figure}[!h]
\centering
\includegraphics[scale=.2, height = 3.5cm]{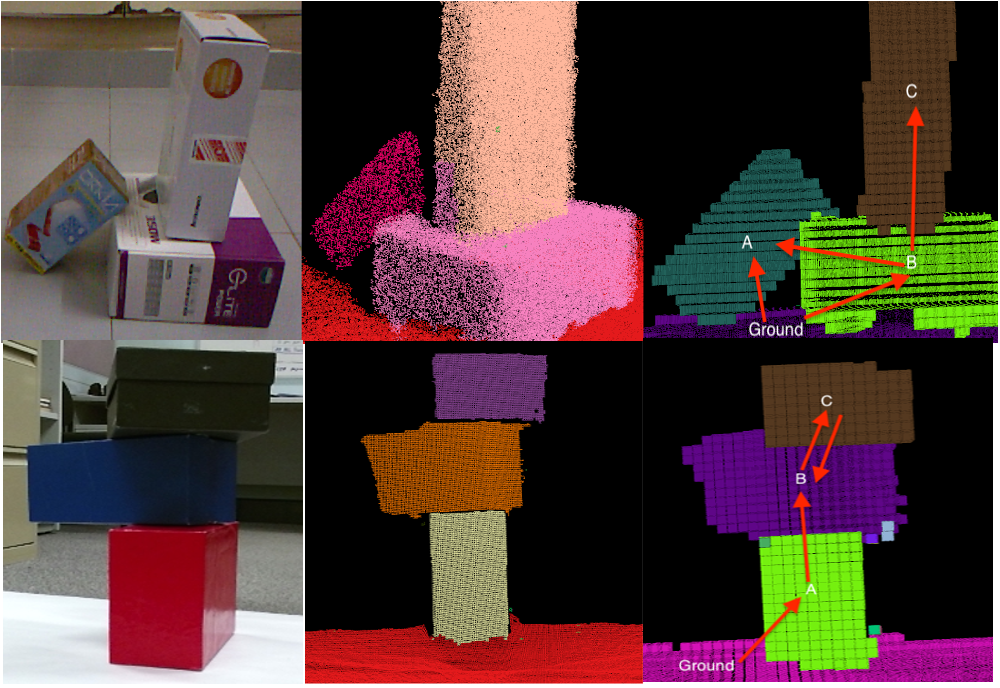}
\captionsetup{justification=centering}
\vspace{-2mm}
\caption{Core supporter detection~\\Top: data set 1;~bottom: single scene.
}
\label{fig:core}
 \vspace{-7mm}
\end{figure}

In addition to data set 1 and 2, we use a single scene with special supporting relations~(e.g. the top object supports the bottom object) as well as the data from \cite{panda2016single}.
Table~\ref{tab:support} shows the core supporter detection results compared with the methods in~\cite{panda2016single}. Our algorithm is able to find most of the true support relations. In addition, our method is able to detect some special core supporter objects such as A in Figure~\ref{fig:core} which is difficult to be detected by statistical methods.

In Figure~\ref{fig:core}, results of core supporter detection are presented. Notably, in the second row, we detected that even though object C is on top of object B, it contributes to the stability of B. Thus, C is a core supporter of B as well as A.

\begin{table}
\scriptsize
\centering
\begin{tabular}{|c|c|c|c|}
\hline
	&Accuracy(Data set 1)&Accuracy(Data set 2)\\
\hline
Our Method&72.5&68.2\\
\hline
Agnostic~\cite{panda2016single}&65.0&N/A\\
\hline
Aware~\cite{panda2016single}&59.5&N/A\\
\hline
\end{tabular}
 \vspace{-1mm}
\caption{Support relation results}
\label{tab:support}
 \vspace{-6mm}
\end{table}

\vspace{-2mm}

\section{Conclusion and Future Work}
In this paper, we propose a framework for identifying support relations among a group of connected objects taking a set of RGB-D images about the same static scene from different views as input.
We assume no knowledge about the objects and the environment beforehand. By qualitatively reasoning about the angle change between each pair of input images, we successfully identified matching of the objects between different views and calculate the initial guess for ICP algorithm.
We use static equilibrium to analyse the stability of the whole structure and extract the core support relation between objects in the structure. We can successfully detect most of the support relations.
With the capability of analysing core supporting relations, the perception system is able to assist the AI agent to perform causal reasoning about consequences of an action applied on an object in a structure. Apparently this is only one aspect of physical relations that can be derived. In the future, more object features~(e.g. solidity, density distribution, etc.) and relations between objects~(e.g. containment, relative position, etc.) can be studied. 


%
\bibliographystyle{named}
{\footnotesize
\bibliography{ijcai19}
}

\end{document}